\documentclass[10pt,twocolumn,letterpaper]{article}

\usepackage{cvpr}
\usepackage{times}
\usepackage{epsfig}
\usepackage{graphicx}
\usepackage{graphbox}
\usepackage{amsmath}
\usepackage{amssymb}

\usepackage{cite}
\usepackage{wrapfig}
\usepackage{multirow}
\usepackage{caption}
\usepackage{subcaption}
\usepackage{booktabs}
\usepackage[ruled,vlined,linesnumbered]{algorithm2e}
\usepackage{comment}
\usepackage{capt-of}
\usepackage{arydshln}

\usepackage[pagebackref=true,breaklinks=true,letterpaper=true,colorlinks,bookmarks=false]{hyperref}

\def\cvprPaperID{514} 

\usepackage{mathtools}

\DeclareMathOperator*{\argmin}{arg\,min}

\newcommand{\norm}[1]{\left\lVert#1\right\rVert}

\newcommand{\proxyS}{\mathbf{\tilde{S}}}
\renewcommand{\S}{\mathbf{S}}

\newcommand{\proxyA}{\mathbf{\tilde{A}}}
\newcommand{\A}{\mathbf{A}}

\newcommand{\Lres}{\mathcal{L}_{\text{res}}}
\newcommand{\Lrecon}{\mathcal{L}_{\text{rec}}}
\newcommand{\Lsym}{\mathcal{L}_{\text{sym}}}
\newcommand{\Lconst}{\mathcal{L}_{\text{con}}}
\newcommand{\Lsmooth}{\mathcal{L}_{\text{smo}}}
\newcommand{\Lreg}{\mathcal{L}_{\text{reg}}}
\newcommand{\Lland}{\mathcal{L}_{\text{lan}}}

\newcommand{\lamconst}{\lambda_{\text{con}}}
\newcommand{\lamsmooth}{\lambda_{\text{smo}}}

\newcommand{\Paragraph}[1]{\vspace{-0mm} \noindent \textbf{#1.} \hspace{0mm}}
\newcommand{\Section}[1]{\vspace{-1mm} \section{#1} \vspace{-1mm}}
\newcommand{\SubSection}[1]{\vspace{-1mm} \subsection{#1} \vspace{-1mm}}

\def\eqnvspace{{\vspace{-2mm}}}
\def\figvspace{{\vspace{-5mm}}}

\cvprfinalcopy 

\ifcvprfinal\fi
\begin{document}

\title{Towards High-fidelity Nonlinear 3D Face Morphable Model}

\author{Luan Tran, Feng Liu, Xiaoming Liu \\
Department of Computer Science and Engineering \\
Michigan State University, East Lansing MI 48824\\
{\tt \{tranluan, liufeng6, liuxm\}@msu.edu}
}

\makeatletter
\let\@oldmaketitle\@maketitle
\renewcommand{\@maketitle}{\@oldmaketitle
\vspace{-6mm}
\begin{center}
\includegraphics[width=0.75\linewidth]{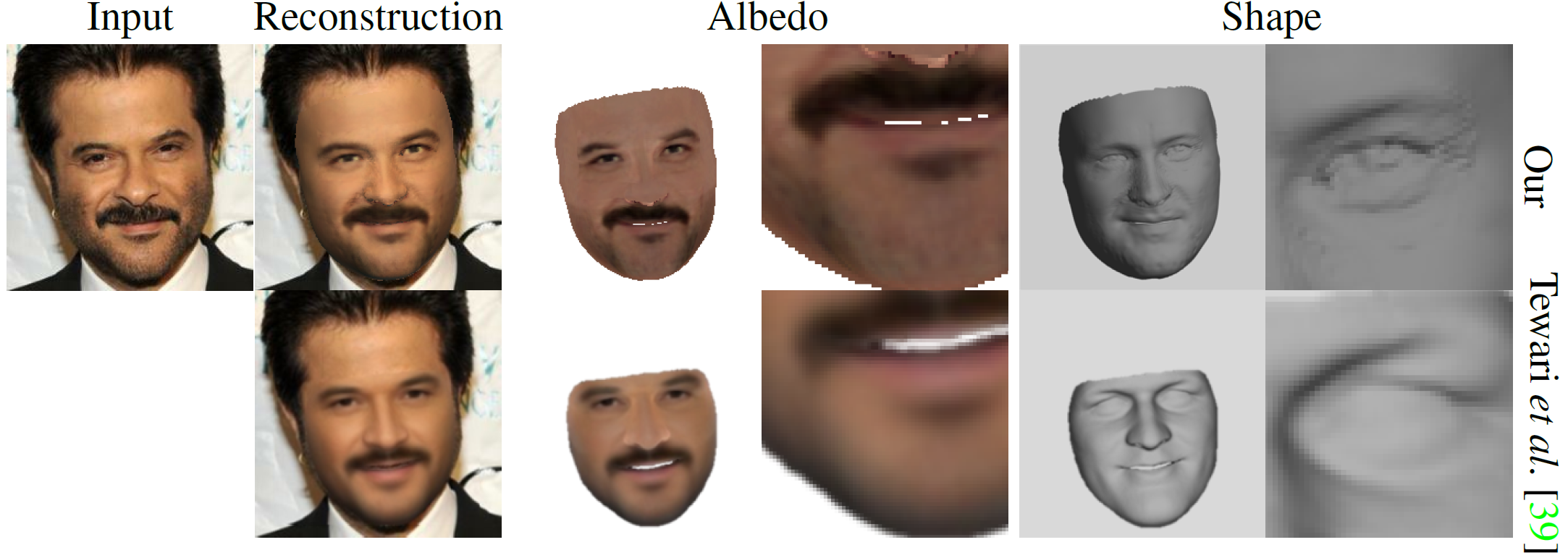}
\vspace{-3mm}
\captionof{figure}{With novel enhancements in both learning objective as well as the network architecture, our proposed nonlinear $3$D morphable model enables, for the first time, regressing high-fidelity facial shape (geometry) and albedo (skin reflectence) by directly estimating model latent representations.}
	\label{fig:teasar}
	
\end{center}
	
    \bigskip}
\makeatother

\maketitle

\begin{abstract}
Embedding $\mathit{3}$D morphable basis functions into deep neural networks opens great potential for 
models with better representation power. However, to faithfully learn those models from an image collection, it requires strong regularization to overcome ambiguities involved in the learning process. This critically prevents us from learning high fidelity face models which are needed to represent face images in high level of details.
To address this problem, this paper presents a novel approach to learn additional proxies as means to side-step strong regularizations, as well as, leverages to promote detailed shape/albedo.
To ease the learning, we also propose to use a dual-pathway network, a carefully-designed architecture that brings a balance between global and local-based models.
By improving the nonlinear $\mathit{3}$D morphable model in both learning objective and network architecture, we present a model which is superior in capturing higher level of details than the linear or its precedent nonlinear counterparts.
As a result, our model achieves state-of-the-art performance on $\mathit{3}$D face reconstruction by solely optimizing latent representations.

Project website:~\url{http://cvlab.cse.msu.edu/project-nonlinear-3dmm.html}
\end{abstract}

\Section{Introduction}
\label{sec:intro}
Computer vision and computer graphics fields have had much interest in the longstanding problem of $3$D face reconstruction --- creating a detailed $3$D model of a person's face from a collection or a single photograph. 
The problem is important with many applications, including but not limited to face recognition~\cite{amberg2008expression,masi2016we,yin2017towards}, video editing~\cite{garrido2015vdub,thies2016face2face}, avatar puppeteering~\cite{bouaziz2013online,cao2014displaced, zell2017facial} or virtual make-up~\cite{garrido2013reconstructing, li2015simulating}.

Recently, an incredible amount of attention is drawn into the simplest but most challenging form of this problem: monocular face reconstruction. 
Inferring a $3$D face mesh from a single $2$D photo is arduous and ill-posed since the image formation process blends multiple facial components (shape, albedo) as well as environment (lighting) into a single color for each pixel.
To better handle the ambiguity, one must rely on additional prior assumptions, such as constraining faces to lie in a restricted subspace, e.g., $3$D Morphable Models ($3$DMM)~\cite{blanz1999morphable} learned from a small $3$D scans collection.
Many state-of-the-art approaches, either learning-based~\cite{richardson20163d, richardson2017learning} or optimization-based face reconstruction~\cite{blanz2003reanimating,garrido2016reconstruction}, heavily rely on such priors.
While yielding impressive results, these algorithms do not generalize well beyond the underlying model's restricted low-dimensional subspace. As a consequence, the reconstructed $3$D face may fail to recover important facial features, contain incorrect details or not well aligned to the input face.

Recently, with the flourishing in neural network, a few attempts have tried to use deep neural networks to replace the $3$DMM basis functions~\cite{tran2018nonlinear,tewari2018self}. 
This increases the model representation power and learns model directly from unconstrained $2$D images to better capture in-the-wild variations. 
However, even with better representation powers, these models still rely on many constraints~\cite{tewari2018self} to regularize the model learning. 
Hence, their objectives involve the conflicting requirements of a strong regularization for a global shape vs.~a weak regularization for capturing higher level details. 
E.g., in order to faithfully separate shading and albedo, albedo is usually assumed to be piecewise constant~\cite{land1971lightness,shu2017neural}, which prevents learning albedo with high level of details.
In this work, besides learning the shape and albedo, we propose to learn additional shape and albedo proxies, on which we can enforce regularizations. 
This also allows us to flexibly pair the true shape with strongly regularized albedo proxy to learn the detailed shape or vice versa. 
As a result, each element can be learned with high fidelity without sacrificing the other element's quality.

On a different note, many $3$DMM models fail to represent small details because of their parameterization.
Many global $3$D face parameterizations have been proposed to overcome the ambiguities associated with single image face fitting such as noise or occlusion. However, because they are designed to model the whole face at once, it is challenging to use them to represent small details. Meanwhile, local-based models can be more expressive than global approaches but with the cost of being less constrained to realistically represent human faces. We propose using dual-pathway networks to provide a better balance between global and local-based models. From the latent space, there is a global pathway focusing on the inference of global face structure and multiple local pathways generating details of different semantic facial parts. Their corresponding features are then fused together for successive process generation of the final shape and albedo. This network also helps to specialize filters in local pathways for each facial part which both improves the quality and saves the computation power.

In this paper, we improve the nonlinear $3$D face morphable model in both learning objective and architecture:
\begin{itemize}
\setlength\itemsep{0em}

\item We solve the conflicting objective problem by learning shape and albedo proxies with proper regularization.

\item The novel pairing scheme allows learning both detailed shape and albedo without sacrificing one.

\item The global-local-based network architecture offers more balance between robustness and flexibility.

\item Our model allows high-fidelity $3$D face reconstruction by solely optimizing latent representations.

\end{itemize}

\Section{Prior Work}
\label{sec:prior}

\Paragraph{Linear $\mathbf{3}$DMM}
The first generic $3$D face model is built by Blanz and Vetter~\cite{blanz1999morphable} using principal component analysis (PCA) on $3$D scans.
%
%
Since this seminal work, there has been a large amount of effort on improving $3$DMM
modeling mechanism.
Paysan~\etal~\cite{paysan20093d} replace the previous UV space alignment~\cite{blanz1999morphable} by Nonrigid Iterative Closest Point~\cite{amberg2007optimal} to directly align $3$D scans.
Vlasic~\etal~\cite{vlasic2005face} use a multilinear model to describe the combined effect of expression and identity variation on the facial geometry.
On the texture side, Booth~\etal~\cite{booth20173d} explore feature-based texture model to represent in-the-wild texture variations.

\Paragraph{Nonlinear face model}
Recently, there is a great interest to use deep neural networks to present the $3$DMM. 
Early work by Duong~\etal~\cite{nhan2015beyond} use Deep Boltzmann Machines to present $2$D Active Appearance Models.
Bagautdinov~\etal~\cite{bagautdinov2018modeling} use Variational Autoencoder (VAE) to learn to model facial geometry directly from $3$D scans.
On another direction, Tewari~\etal~\cite{tewari2018self} and Tran and Liu~\cite{tran2018nonlinear} attempt to learn $3$DMM models from a $2$D image collection.
Tewari~\etal~\cite{tewari2018self} embed shape and albedo bases in multi-layer perceptions.
Meanwhile, Tran and Liu~\cite{tran2018nonlinear} use convolution neural networks by representing both geometry and skin reflectance in UV space.
Despite having greater representation power, these models still have difficulty in recovering small details in the input images due to strong regularizations in their learning objectives. 

\Paragraph{Global/local-based facial parameterization}
Although, global $3$D face parameterizations\cite{lau2009face,vlasic2005face} can remedy the vagueness associated with monocular face tracking~\cite{black1995tracking, decarlo1996integration};
they can't represent small geometry details without making them exceedingly large and unwieldy.
Hence, region or local-based models are proposed to overcome this problem.
Blanz and Vetter~\cite{blanz1999morphable} and Tena~\etal~\cite{tena2011interactive} learn
a region-based PCA, where Blanz and Vetter~\cite{blanz1999morphable} segment the face into semantic subregions (eyes, nose, mouth), while Tena~\etal~\cite{tena2011interactive} further split into smaller regions to increase the model's expressiveness. 
Other approaches include a region-based blendshape~\cite{joshi2006learning} or localized multilinear model~\cite{brunton2014multilinear}.
All these models bring more flexibility than the global one but at the cost of being less constrained on realistically representing human faces.
Our approach offers a balance between global and local models by using a dual-pathway network architecture.
Bagautdinov~\etal~\cite{bagautdinov2018modeling} try to achieve a similar objective with compositional VAE by introducing multiple layers of hidden variables, but at a cost of extremely large numbers of hidden variables.

\Paragraph{Residual learning}
Residual learning has been used in many vision tasks. In super resolution, Kim~\etal~\cite{kim2016accurate} propose to learn the difference between the high-resolution target and the low-resolution input rather than estimating the target itself.
%
In face alignment~\cite{jourabloo2015pose}, or missing data imputation task~\cite{tran2017missing}, residual learning is used in many cascade of networks to iteratively refine their estimation by learning the difference with the true target. 
In this work, we leverage residual learning idea but with a different purpose to overcome conflicting objectives in learning $3$D models.

\begin{figure*}[t!]
\centering
\includegraphics[trim=0 0 0 0,clip, width=0.8\linewidth]{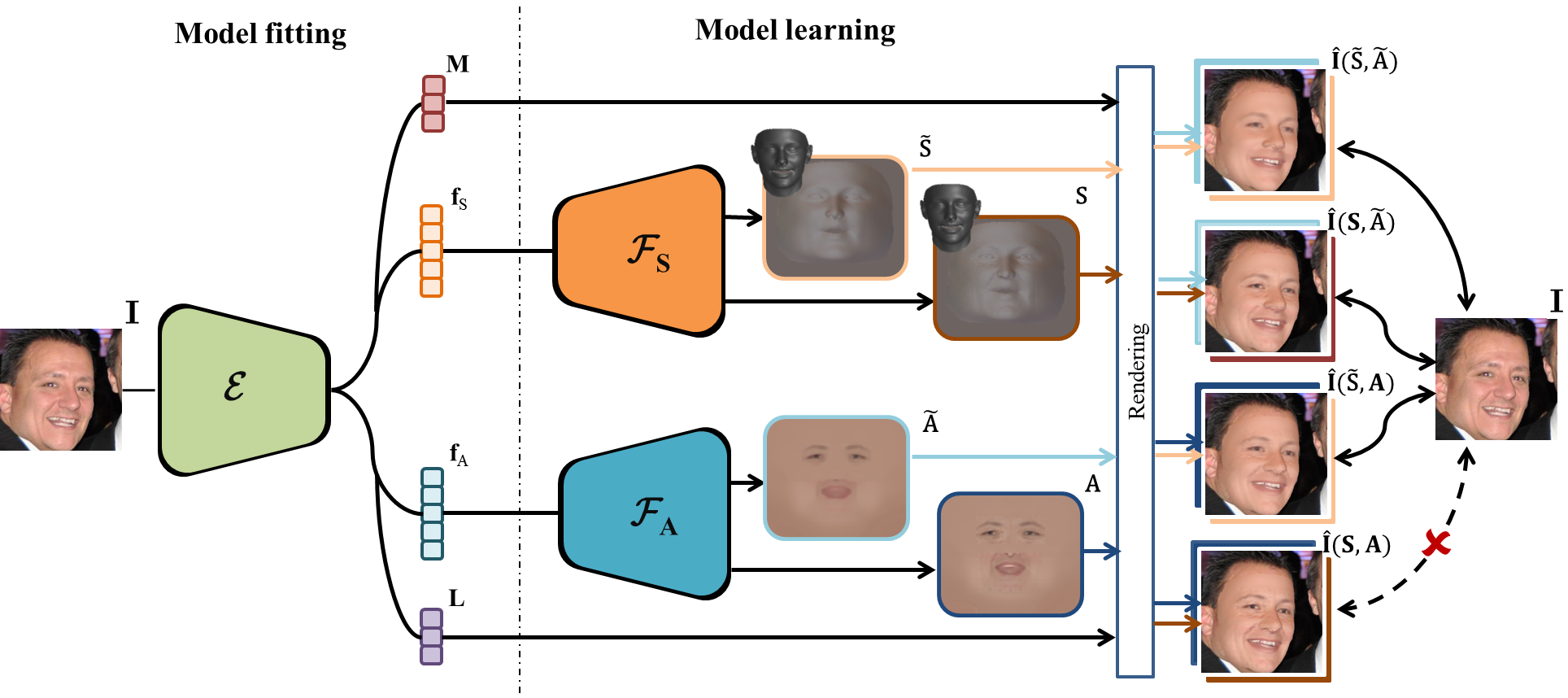}
\caption{\small The proposed framework. Each shape or albedo decoder consist of two branches to reconstruct the true element and its proxy. Proxies free shape and albedo from strong regularizations, allow them to learn models with high level of details. }
\label{fig:architecture}
\end{figure*}

\Section{Proposed Method}
\label{sec:alg}

For completeness, we start by briefly summarizing  the traditional linear $3$DMM, the recently proposed nonlinear $3$DMM learning method including their limitations. Then we introduce our proposed improvements in both learning objective and network architecture.

\SubSection{Linear $\mathbf{3}$DMM}
\label{sec:linear}
The $3$D Morphable Model ($3$DMM)~\cite{blanz1999morphable} provides parametric models representing faces using two components: shape (geometry) and albedo (skin reflectance).
Blanz~\etal~\cite{blanz1999morphable} describe the $3$D face space with PCA. The $3$D face mesh $\S\in\mathbb{R}^{3Q}$ with $Q$ vertices is computed as:
\begin{equation}
\S = \mathcal{F}_S(\mathbf{f}_S | \mathbf{\Theta}_S) = \mathbf{\Theta}_S \mathbf{f}_S,
\eqnvspace
\end{equation}
where $\mathcal{F}_S( \mathbf{f}_S | \mathbf{\Theta}_S)$ is a function of $\mathbf{f}_S\in\mathbb{R}^{l_S}$, parameterized by $\mathbf{\Theta}_S$. In linear model, $\mathcal{F}_S$ is simply a matrix multiplication (the mean shape is omitted for clarity).

The albedo of the face $\mathbf{A}\in\mathbb{R}^{3Q}$ is defined within a template shape, describing the R, G, B colors of $Q$ corresponding vertices.
$\mathbf{A}$ is also formulated in a similar fashion:
\begin{equation}
\A = \mathcal{F}_A(\mathbf{f}_A | \mathbf{\Theta}_A) = \mathbf{\Theta}_A \mathbf{f}_A.
\eqnvspace
\end{equation}


To synthesize $2$D face images, the $3$D mesh is projected onto the image plan with the weak perspective projection model.
Then, the texture and $2$D image is rendered using an illumination model, i.e.,  Spherical Harmonics~\cite{ramamoorthi2001efficient}.

\SubSection{Nonlinear $\mathbf{3}$DMM}
\label{sec:nonliear3dmm}

Recently, Tewari~\etal~\cite{tewari2018self}, Tran and Liu~\cite{tran2018nonlinear,tran2018on} concurrently propose to use deep neural network to present $3$DMM bases. Essentially, mappings $\mathcal{F}_S$ and $\mathcal{F}_A$ are now represented as neural networks with parameters $\Theta_S, \Theta_A$ respectively. Tewari~\etal~\cite{tewari2018self} straightforwardly use multi-layer perceptron as their networks. Meanwhile, Tran and Liu~\cite{tran2018nonlinear} leverage spatial relation of vertices by presenting both $\S$ and $\A$ in a UV space, denoted $\S^{\text{UV}}, \A^{\text{UV}}$. Mappings $\mathcal{F}_{*}$ are convolution neural networks (CNNs) with an extra sampling step converting from $\mathbb{R}^{\text{UV}}$ to $\mathbb{R}^{3Q}$.
To make the framework end-to-end trainable, they also learn a model fitting module, $\mathcal{E}$, which is another CNN.
Beside estimating shape, albedo latent vectors $\mathbf{f}_S, \mathbf{f}_A$, the encoder $\mathcal{E}$ also estimates projection matrix $\mathbf{M}$ as well as lighting coefficients $\mathbf{L}$. 
The objective of the whole network is to reconstruct the original input image via a differentiable rendering layer $\mathcal{R}$:
\begin{gather}
\argmin_{\mathcal{E},\mathcal{F}_S, \mathcal{F}_A} \sum_{ \mathbf{I}} \Lrecon(\hat{\mathbf{I}} , \mathbf{I}), \\
\hat{\mathbf{I}} = \mathcal{R} \left( \mathcal{E}_M(\mathbf{I}), \mathcal{E}_L(\mathbf{I}), \mathcal{F}_S(\mathcal{E}_S(\mathbf{I})), \mathcal{F}_A(\mathcal{E}_A(\mathbf{I})) \right). \nonumber
\end{gather}

\Paragraph{Reconstruction loss}
There are many design options for the reconstruction loss. The straightforward choice is comparing images in the pixel space, with typical $l_1$ or $l_2$ loss. To better handle outliers, the robust $l_{2,1}$ is adopted, where the distance in the RGB color space is based on $l_2$ and the summation is based on $l_1$-norm to enforce sparsity~\cite{thies2016face2face,thies2016facevr}:
\begin{equation}
\mathcal{L}^i_{\text{rec}} = \frac{1}{|\mathcal{V}|} \sum_{q \in \mathcal{V} }\norm{\mathbf{\hat{I}}(q) - \mathbf{I}(q)}_2,
\label{eqn:reconLoss}
\eqnvspace
\end{equation}
where $\mathcal{V}$ is the set of pixels covered by the estimated mesh.

The closeness between images $\mathbf{\hat{I}}$ and $\mathbf{I}$ can also be enforced in the feature space (perceptual loss):
\begin{equation}
\mathcal{L}^f_{\text{rec}} = \frac{1}{|\mathcal{C}|} \sum_{j \in \mathcal{C}} \frac{1}{W_jH_jC_j}|| \varphi_j(\mathbf{\hat{I}}) - \varphi_j(\mathbf{I})||^2_2.
\label{eqn:perceptualLoss}
\eqnvspace
\end{equation}
The loss is summed over $\mathcal{C}$, a subset of layers of the network $\varphi$. Here $\varphi_j(\mathbf{I})$ is the activations of the $j$-th layer of $\varphi$ with dimension $W_j \times H_j \times C_j$ obtained when processing $\mathbf{I}$.
%


The final reconstruction loss is a weighted average between the image and feature reconstruction losses:
\begin{equation}
\Lrecon(\hat{\mathbf{I}} , \mathbf{I}) = \Lrecon^i(\hat{\mathbf{I}} , \mathbf{I}) + \lambda_f \Lrecon^f(\hat{\mathbf{I}} , \mathbf{I}).
\end{equation}

\Paragraph{Sparse Landmark Alignment}
To help achieve better model fitting, which in turn helps to improve the model learning itself, the landmark alignment loss is used as an auxiliary task. The loss is defined by Euclidean distance between estimated and groundtruth landmarks:
\begin{align}
\Lland = & \norm {\mathbf{M} \ast \begin{bmatrix} \S(:,\mathbf{d}) \\ \mathbf{1} \end{bmatrix} - \mathbf{U} }^2_2,
\label{eq:landmarkloss}
\eqnvspace
\end{align}
where $\mathbf{U} \in \mathbb{R}^{2{\times} 68}$ is the manual labels of $2$D landmark locations, $\mathbf{d}$ stores the indexes of $68$ vertices corresponding to the sparse $2$D landmarks in the $3$D face mesh.
In ~\cite{tran2018nonlinear,tran2018on}, the landmark loss is only applied on $\mathcal{E}$ to prevent learning implausible shapes as the loss only affects a tiny subsets of vertices related to the keypoints.

\Paragraph{Different regularization}
To overcome ambiguity and faithfully recover different elements (shape, albedo, lighting), many regularizations are needed.

\textit{Albedo Symmetry}:
\begin{equation}
\Lsym(\A) = \norm{ \mathbf{A}^{\text{uv}} - \text{flip}(\A^{\text{uv}}) }_1,
\label{eqn:alb_sym}
\end{equation}

where  $\text{flip}()$ is a horizontal image flip operation.

\textit{Albedo Constancy}:\\
\noindent \resizebox{0.95\linewidth}{!}{
\begin{minipage}{\linewidth}
\begin{eqnarray}
\Lconst(\A) =  \sum_{\mathbf{v}_j^{\text{uv}} \in \mathcal{N}_i} \omega (\mathbf{v}_i^{\text{uv}}, \mathbf{v}_j^{\text{uv}}) \norm{ \A^{\text{uv}}(\mathbf{v}_i^{\text{uv}}) - \A^{\text{uv}}(\mathbf{v}_j^{\text{uv}}) }_2^p.
\label{eqn:alb_const}
\end{eqnarray}
\vspace{0.5mm}
\end{minipage}
}
The weight $\omega (\mathbf{v}_i^{\text{uv}}, \mathbf{v}_j^{\text{uv}}) = \exp\left(-\alpha \norm{ \mathbf{c}(\mathbf{v}_i^{\text{uv}}) - \mathbf{c}(\mathbf{v}_j^{\text{uv}} ) } \right) $, helps to penalize more on pixels with the same chromaticity (i.e., $\mathbf{c}(x) = \mathbf{I}(x) / |\mathbf{I}(x)|$), where the color is referenced from the input image using the current estimated projection. $\mathcal{N}_i$ denotes a set of $4$-pixel neighborhood of pixel $\mathbf{v}_i^{\text{uv}}$.

\textit{Shape Smoothness}: This is a Laplacian regularization on the vertex locations.

\noindent \resizebox{0.96\linewidth}{!}{
\begin{minipage}{\linewidth}
\begin{eqnarray}
\Lsmooth(\S) = \sum_{\mathbf{v}_i^{\text{uv}} \in \S^{\text{uv}}} \norm{ \S^{\text{uv}}(\mathbf{v}_i^{\text{uv}}) - \frac{1}{|\mathcal{N}_i|}  \sum_{\mathbf{v}_j^{\text{uv}} \in \mathcal{N}_i} \S^{\text{uv}}(\mathbf{v}_j^{\text{uv}}) }_2.
\label{eqn:shape_smooth}
\end{eqnarray}
\vspace{0.5mm}
\end{minipage}
}
The overall objective can be summarized  as:
\begin{equation}
\mathcal{L} = \Lrecon(\hat{\mathbf{I}} , \mathbf{I}) + \Lland + \Lreg,
\end{equation}
\begin{equation}
\text{with } \Lreg = \Lsym(\A) + \lamconst\Lconst(\A) + \lamsmooth\Lsmooth(\S).
\end{equation}

\SubSection{Nonlinear $\mathbf{3}$DMM with Proxy and Residual }
\Paragraph{Proxy and Residual Learning}
Strong regularization has been shown to be critical in ensuring the plausibility of the learned models~\cite{tewari2018self, tran2018on}. However, the strong regularization also prevents the model from recovering high-level details in either shape or albedo.
Hence, this prevents us from achieving the ultimate goal of learning a high-fidelity $3$DMM model.

In this work, we propose to learn additional \textbf{proxy shape}~($\proxyS$) and \textbf{proxy albedo}~($\proxyA$), on which we can apply the regularization. All presented regularizations will now be moved to proxies:
\begin{equation}
\Lreg^* = \Lsym(\proxyA) + \lamconst\Lconst(\proxyA) + \lamsmooth\Lsmooth(\proxyS).
\end{equation}

There will be no regularization applied directly to the actual shape $\S$ and albedo $\A$, other than a weak regularization encouraging each to be close to its proxy:

\begin{equation}
\Lres = \norm{\Delta \S}_1 + \norm{\Delta \A}_1 = \norm{\S - \proxyS}_1 + \norm{\A - \proxyA}_1.
\label{eqn:regu_final}
\end{equation}

By pairing two shapes $\S, \proxyS$ and two albedos $\A, \proxyA$, we can render four different output images (Fig.~\ref{fig:architecture}). Any of them can be used to compare with the original input image. We rewrite our reconstruction loss as:
\begin{align}
\Lrecon^*  &= \Lrecon( \hat {\mathbf{I}}(\proxyS,\proxyA), \mathbf{I} ) \nonumber \\
           &+ \Lrecon( \hat {\mathbf{I}}(\proxyS,\A), \mathbf{I} ) \nonumber \\
           &+ \Lrecon( \hat {\mathbf{I}}(\S,\proxyA), \mathbf{I} ).
\label{eqn:recon_final}           
\end{align}

Pairing strongly regularized proxies and weakly regularized components is a critical point in our approach. Using proxies allows us to learn high-fidelity shape and albedo without sacrificing quality of either component. This pairing is inspired by the observation that Shape from Shading techniques are able to recover detailed face mesh by assuming over regularized albedo or even using the mean albedo~\cite{richardson2017learning}. Here, $\Lrecon( \hat {\mathbf{I}}(\S,\proxyA), \mathbf{I} )$ loss promotes $\S$ to recover more details as $\proxyA$ is constrained by piece-wise constant $\Lconst(\proxyA)$ objective. Vice versa, $\Lrecon( \hat {\mathbf{I}}(\proxyS,\A), \mathbf{I} )$ aims to learn better albedo. In order for these two losses to work as desired, proxies $\proxyS$ and $\proxyA$ should perform well enough to approximate the input images by themselves.
%
Without $\Lrecon( \hat {\mathbf{I}}(\proxyS,\proxyA), \mathbf{I} )$, a valid solution that minimizes $\Lrecon( \hat {\mathbf{I}}(\S,\proxyA), \mathbf{I} )$ is combination of a constant albedo proxy and noisy shape creating surface normal with dark shading in necessary regions, i.e., eyebrows.


Another notable design choice is that we intentionally left out the loss function on $\hat {\mathbf{I}}(\S,\A)$, even though this theoretically is the most important objective.
This is to avoid the case that the shape $\S$ learns an in-between solution that works well with both $\proxyA, \A$ and vice versa. 

\Paragraph{Occlusion Imputation}
With proposed objective function, our model is able to faithfully reconstruct input images. However, we empirically found that besides high-fidelity visible regions, the model tends to keep invisible region smooth. The reason might be that, there is no supervision on those areas other than the residual magnitude loss pulling the shape and albedo closer to their proxies.
To learn a more meaningful model, which is beneficial to other applications, i.e., face editing or face synthesis, we propose to use a soft symmetry loss~\cite{tran2017extreme} on occluded regions:
\begin{equation}
\mathcal{L}_{\text{res-sym}}(\S) = \norm{ \mathbf{T} \odot (\Delta \S_z^{\text{uv}} - \text{flip}( \Delta \S_z^{\text{uv}})) }_1,
\label{eqn:alb_sym}
\end{equation}
where $\mathbf{T}$ is a visibility mask of each pixel in UV space, approximated based on estimated surface normal direction.
Even though the shape itself is not symmetric, i.e., face with asymmetric expression; we enforce symmetrical property on its depth residual $\Delta \S_z$ (only use shape's $z$-dimension).

\SubSection{Global-Local-Based Network Architecture}
\label{sec:network}

\begin{figure}[t!]
\centering
\includegraphics[trim=0 5 0 30,clip, width=0.95\linewidth]{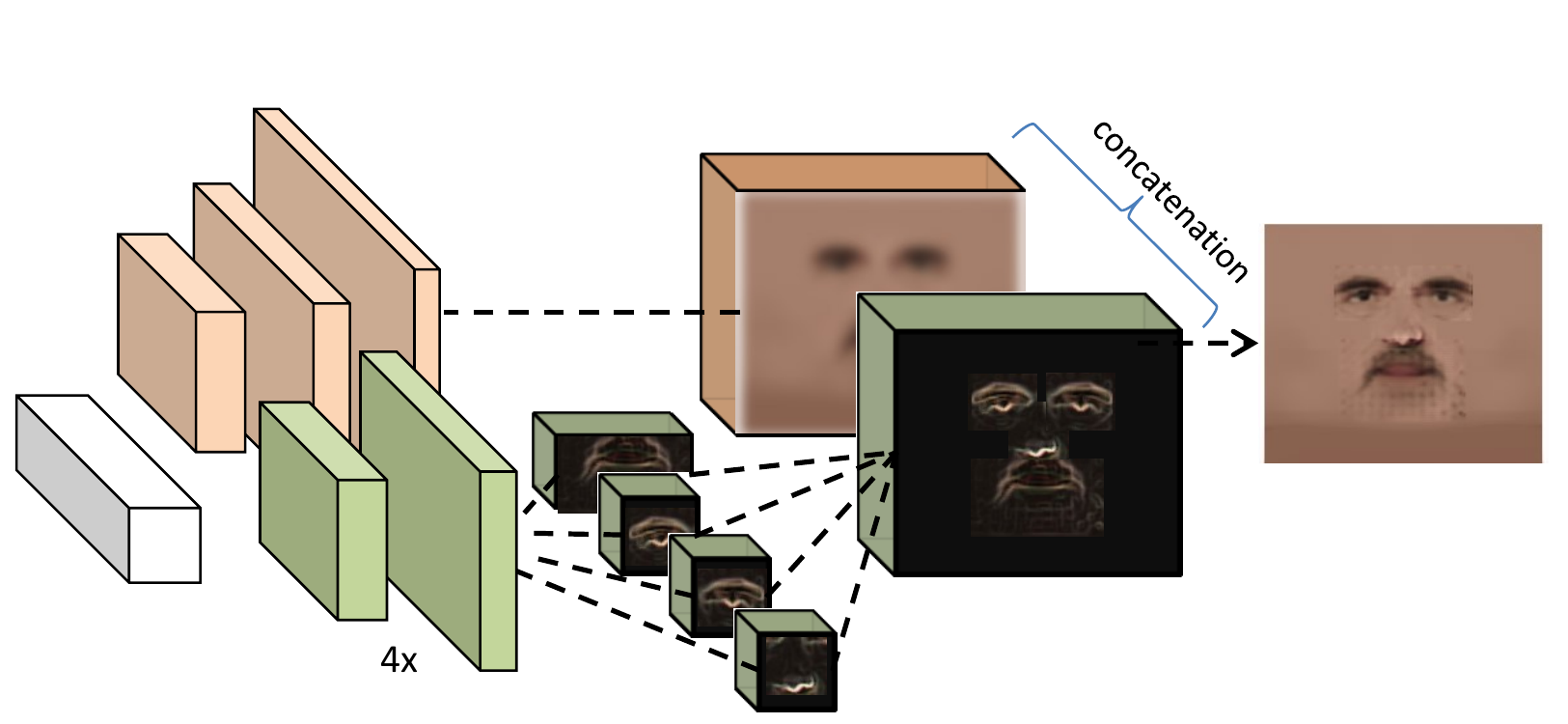}
\vspace{-1mm}
\caption{\small The proposed global-local-based network architecture. }
\label{fig:network}
\vspace{-4mm}
\end{figure}

While global-based models are usually robust to noise and mismatches, they are usually over-constrained and do not provide sufficient flexibility to represent high-frequency deformations as local-based models. In order to take the best of both worlds, we propose to use dual-pathway networks for our shape and albedo decoders.

Here, we transfer the success of combining local and global models in image synthesis~\cite{mohammed2009visio,huang2017beyond} to $3$D face modeling.
The general architecture of a decoder is shown in Fig.~\ref{fig:network}.
From the latent vector, there is a global pathway focusing on inferring the global structure and a local pathway with four small sub-networks generating details of different facial parts, including eyes, nose and mouth.
The global pathway is built from fractional strided convolution layers with five up-sampling steps. Meanwhile, each sub-network in the local pathway has the similar architecture but shallower with only three up-sampling steps. 
%
Using different small sub-networks for each facial part offers two benefits: i) with less up-sampling steps, the network is better able to represent high-frequency details in early layers; ii) each sub-network can learn part-specific filters, which is more computationally efficient than applying across global face.

%
%
As shown in Fig.~\ref{fig:network}, to fuse two pathways' features, we firstly integrate four local pathways' outputs into one single feature tensor.
%
Different from other works that synthesize face images with different yaw angles ~\cite{tran2017disentangled,tran2018representation,karras2017progressive} with no fixed keypoints' locations, our $3$DMM generates facial albedo as well as $3$D shape in UV space with predefined topology. 
Merging these local feature tensors is efficiently done with the zero padding operation.
The max-pooling fusion strategy is also used to reduce the stitching artifacts on the overlapping areas.
Then the resultant feature is simply concatenated with the global pathway's feature, which has the same spatial resolution. 
Successive convolution layers integrate information from both pathways and generate the final albedo/shape (or their proxies). 
\Section{Experimental Results}
\label{sec:exp}

We study different aspects of the proposed framework, in terms of framework design, model representation power, and applications to facial analysis.

The training is similar to ~\cite{tran2018on}, which also include a pretrain stage with supervised losses. Adopting Basel Face Model~(BFM)~\cite{paysan20093d}'s facial mesh triangle topology, we use a subset of $Q=39,111$ vertices on the face region only. The model is trained on 300W-LP dataset~\cite{zhu2016face}, which contains $122,450$ in-the-wild face images, in a wide pose range.

The model is optimized using Adam optimizer with a learning rate of $0.001$. We set the following parameters: $U=192, V=224$, $l_S=l_A=320$. $\lambda$~values are selected to bring losses to similar magnitudes.

\SubSection{Ablation Study}

\begin{figure}[t!]
\begin{center}
\includegraphics[width=0.99\linewidth]{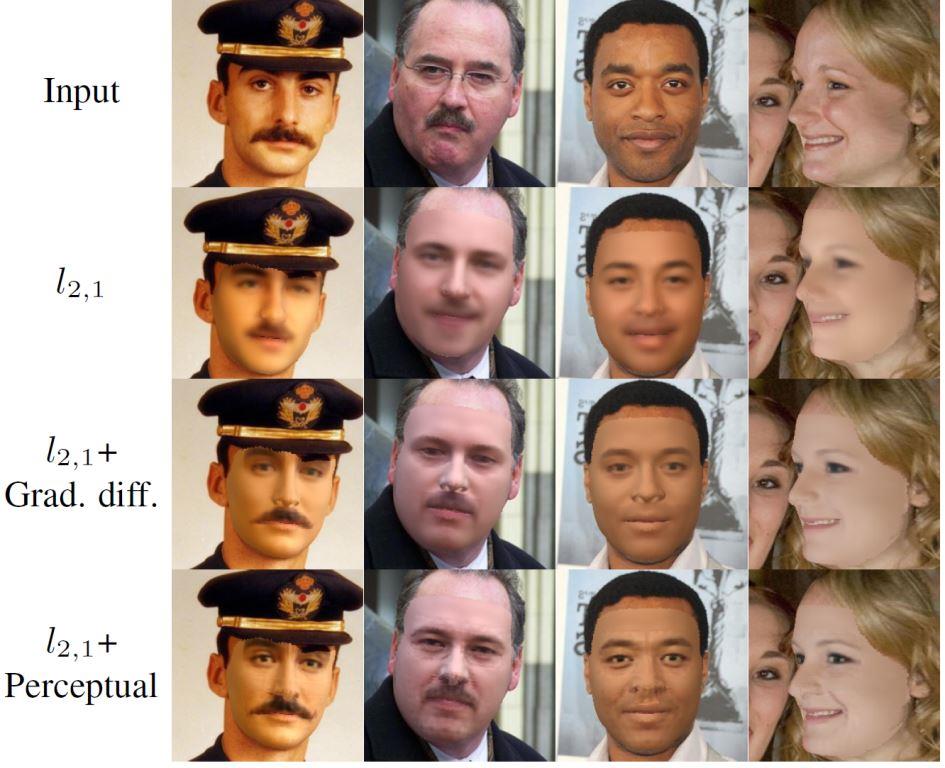}
\vspace{-2mm}
\caption{\small Reconstruction results with different loss functions.}
\label{fig:recon_loss}
\figvspace
\end{center}
\end{figure}

\Paragraph{Reconstruction Loss Functions}
We study effects of different reconstruction losses on quality of the reconstructed images~(Fig.~\ref{fig:recon_loss}).
As expected, the model trained with $l_{2,1}$ loss only results in blurry reconstruction, similar to other $l_p$ loss. 
To make the reconstruction more realistic, we explore other options such as gradient difference~\cite{mathieu2015deep} or perceptual loss~\cite{johnson2016perceptual}.
While adding the gradient difference loss creates more details in the reconstruction, combining perceptual loss with $l_{2,1}$ gives the best results with high level of details and realism. 
For the rest of the paper we will refer to the model trained using this combination.

\begin{figure}[t!]
\begin{center}
\includegraphics[width=0.99\linewidth]{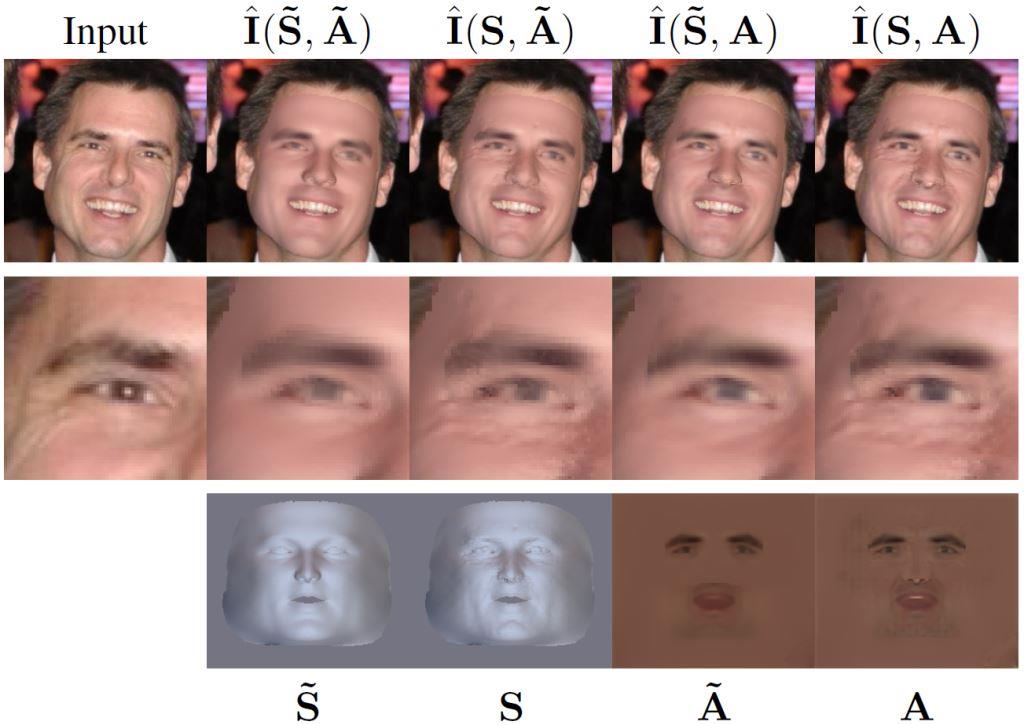}
\vspace{-2mm}
\caption{\small Image reconstruction with our $3$DMM model using the proxy and the true shape and albedo. Our shape and albedo can faithfully recover details of the face. Note: for the shape, we show the shading in UV space -- a better visualization than the raw $\S^{\text{UV}}$.}
\label{fig:different_images}
\figvspace 
\end{center}
\end{figure}

\def\SoftSymHeight{0.087\textwidth}
\begin{figure}[t!]
\begin{center}
\includegraphics[width=0.99\linewidth]{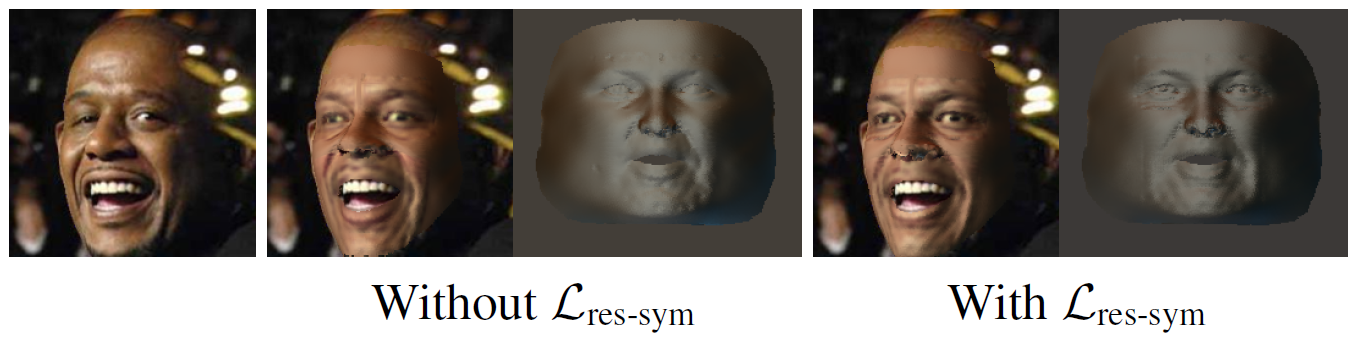}
\vspace{-2mm}
\caption{\small Affect of soft symmetry loss on our shape model. }
\label{fig:soft_sym}\figvspace 
\end{center}
\end{figure}

\Paragraph{Understanding image pairing}
Fig.~\ref{fig:different_images} shows fitting results of our model on a $2$D face image. By using the proxy or the final components (shape or albedo) we can render four different reconstructed images with different quality and characteristics. The image generated by two proxies $\proxyS, \proxyA$ is quite blurry but is still be able to capture major variations in the input face. By pairing $\S$ and the proxy $\proxyA$, $S$ is enforced to capture high level of details to bring the image closer to the input. Similarly, $\A$ is also encouraged to capture more details by pairing with the proxy $\proxyS$. The final image $\hat{\mathbf{I}}(\S,\A)$ inherently achieves high level of details and realism even without direct optimization.

\Paragraph{Residual Soft Symmetry Loss}
We study effects of the residual soft symmetry loss on recovering details on occluded face region. 
As shown in Fig.~\ref{fig:soft_sym}, without $\mathcal{L}_{\text{res-sym}}$, the learned model can result in an unnatural shape, in which one side of the face is over-smooth, on occluded regions, while the other side still has high level of details. 
Our model learned with $\mathcal{L}_{\text{res-sym}}$ can consistently create details across the face, even in occluded areas.

\SubSection{Representation Power}
\label{sec:representation_power}
We compare the representation power of the proposed nonlinear $3$DMM with Basel Face Model~\cite{paysan20093d}, the most commonly used linear $3$DMM. We also make comparisons with the recently proposed nonlinear $3$DMM~\cite{tran2018nonlinear}.

\begin{figure}[t!]
\begin{center}
\small
\includegraphics[width=0.99\linewidth]{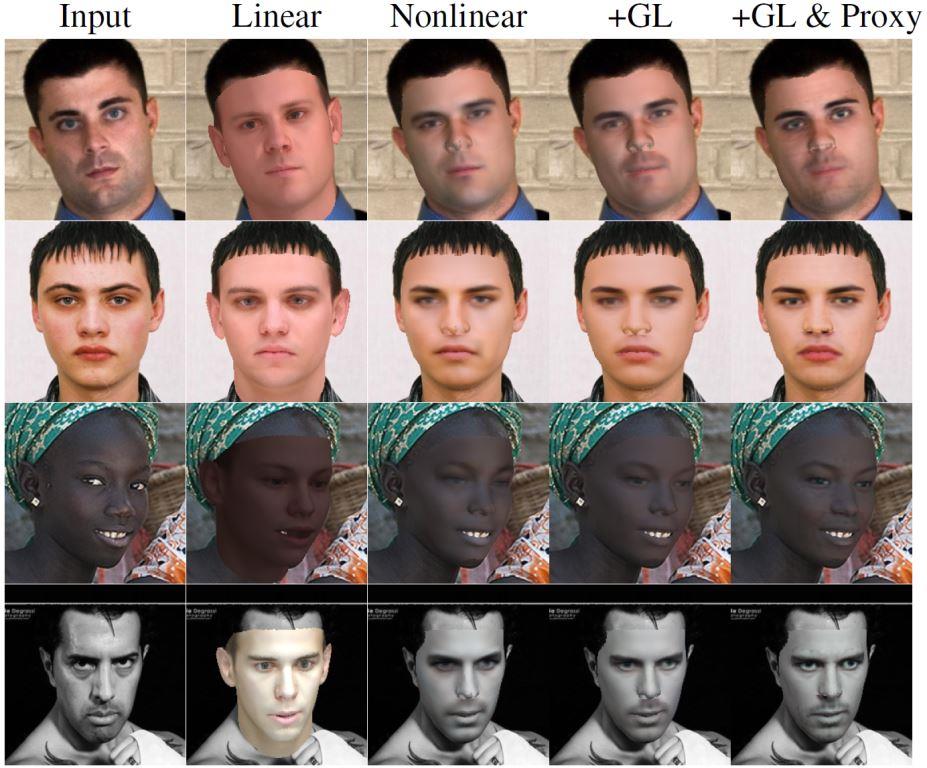}
\vspace{-3mm}
\caption{\small Qualitative comparisons on texture representation power. Our model can better reconstruct in-the-wild facial texture.}
\label{fig:tex_representationpower}
\figvspace
\end{center}
\end{figure}

\begin{table}[t!]
\footnotesize
\caption{\small{Texture representation power quantitative comparison (Average reconstruction error on non-occluded face portion.)}} 
\label{tab:tex_representation_tab}
\figvspace
\begin{center}
\begin{tabular}{ lc}
\toprule 
Method & Reconstruction error ($l_{2,1}$) \\ \midrule
Linear~\cite{zhu2016face} & $0.1287$\\
Nonlinear~\cite{tran2018on} & $0.0427$ \\
Nonlinear + GL (Ours) & $0.0386$ \\ 
Nonlinear + GL + Proxy (Ours) & $\mathbf{0.0363}$   \\ 
\bottomrule
\end{tabular}
\end{center}
\figvspace
\end{table}

\Paragraph{Texture}
We evaluate our model's power to represent in-the-wild facial texture on AFLW2000-3D dataset~\cite{zhu2016face}. Given a face image, also with the groundtruth geometry and camera projection, we can jointly estimate an albedo parameter $\mathbf{f}_A$ and a lighting parameter $\mathbf{L}$ whose decoded texture can reconstruct the original image. To accomplish this, we use SGD on $\mathbf{f}_A$ and $\mathbf{L}$ with the initial parameters estimated by our encoder $\mathcal{E}$. 
For the linear model, Zhu~\etal~\cite{zhu2016face} fitting results of Basel albedo using Phong illumination model~\cite{phong1975illumination} is used. 
%
As in Fig.~\ref{fig:tex_representationpower}, nonlinear model significantly outperforms the Basel Face model. 
Despite, being close to the original image, Tran and Liu~\cite{tran2018on} model reconstruction results are still blurry.
Using global-local-based network architecture (``+GL'') with the same loss functions helps to bring the image closer to the input.
However, these models are still constrained by regularizations on the albedo. By learning using proxy technique (``+Proxy''), our model can learn more realistic albedo with more high frequency details on the face.
This conclusion is further supported with quantitative comparison in Tab.~\ref{tab:tex_representation_tab}.
We report the averaged $l_{2,1}$ reconstruction error over the face portion of each image. Our model achieves the lowest averaged reconstruction error among four models, $0.0363$, which is a $15\%$ error reduction of the recent nonlinear $3$DMM work~\cite{tran2018on}.

\begin{figure}[t!]
\begin{center}
\includegraphics[width=0.9\linewidth]{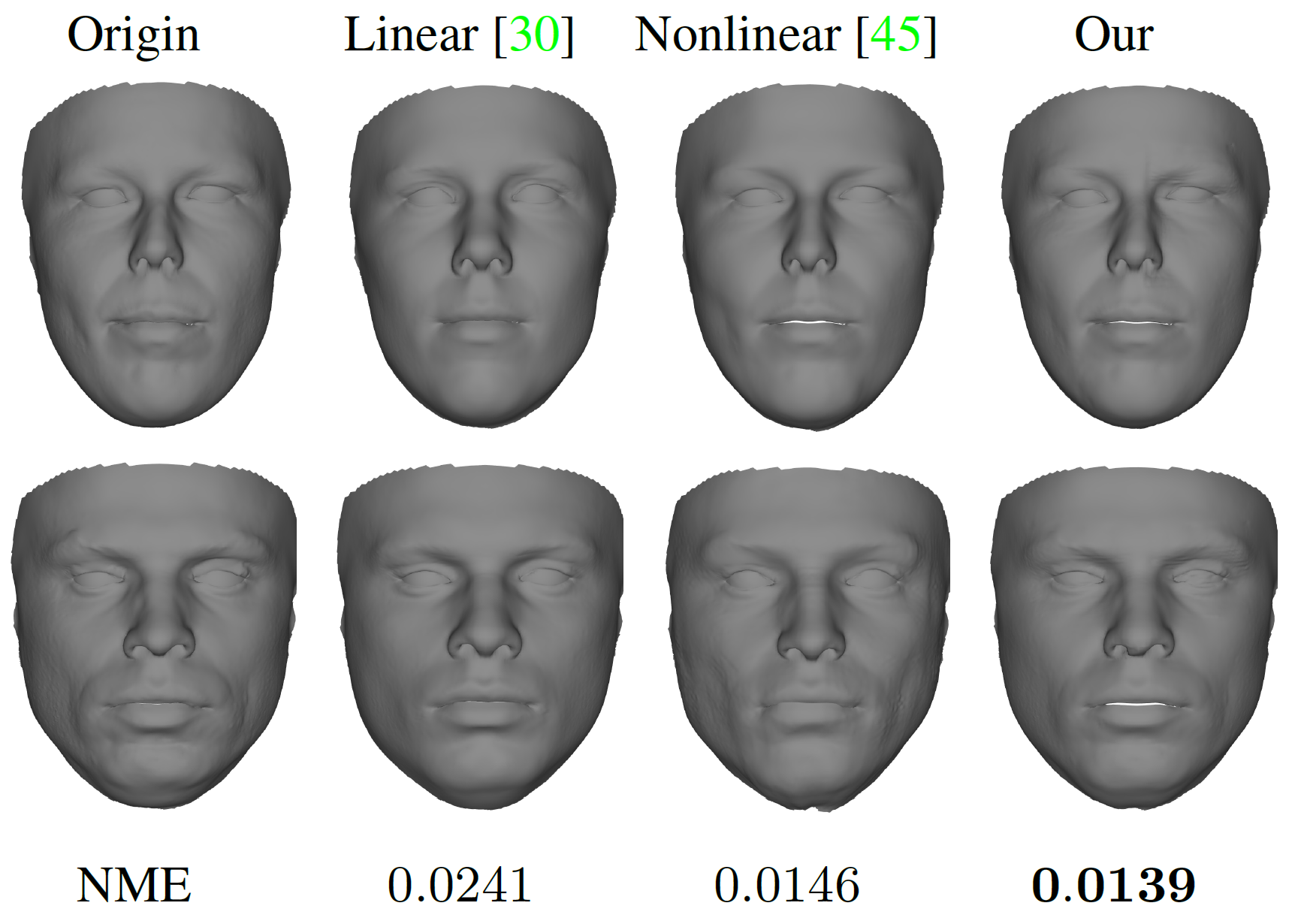}
\vspace{-2mm}
\caption{\small Shape representation power comparison. Given a $3$D shape, we optimize the feature $\mathbf{f}_S$ to approximate the original one. }
\label{fig:shape_representation}
\figvspace
\end{center}
\end{figure}

\begin{figure}[t!]
\centering
\includegraphics[trim=0 0 0 0,clip, width=0.99\linewidth]{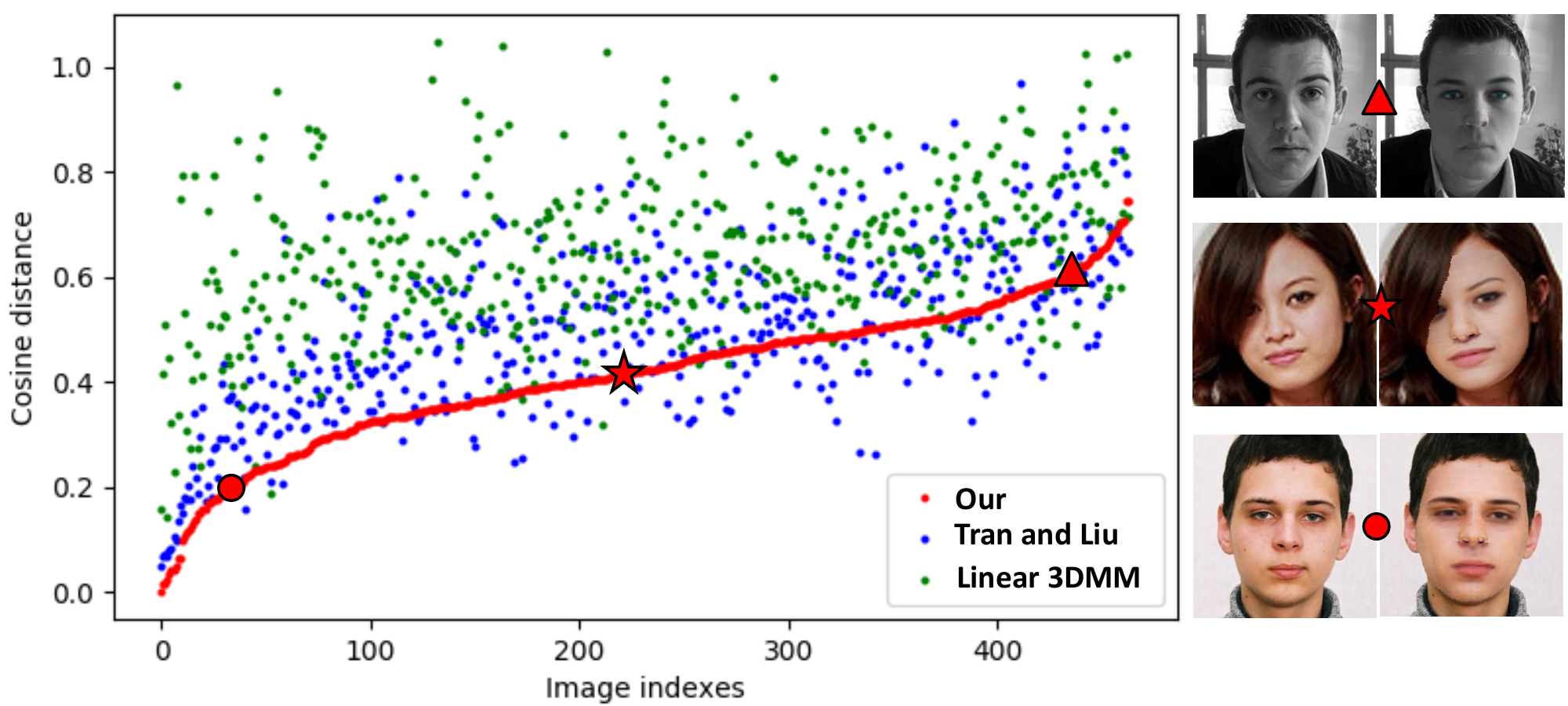}
\caption{\small The distance between the input images and their reconstruction from three models. For better visualization, images are sorted based on their distance to our model's reconstructions.}
\label{fig:id_preserving}\figvspace
\end{figure}

\Paragraph{Shape}
Similarly, we also compare models' power to represent real-world $3$D scans.
Using ten $3$D face meshes provided by~\cite{paysan20093d}, which share the same triangle topology with us, we can optimize the shape parameter to generate, through the decoder, shapes matching the groundtruth scans.
The optimization objective is defined based on vertex distances (Euclidean) as well as surface normal direction (cosine distance), which empirically improves reconstructed meshes' fidelity compared to optimizing the former only.
Fig.~\ref{fig:shape_representation} shows the visual comparisons between different reconstructed meshes. Our reconstructions closely match the face shapes details. 
%
To make quantitative comparisons, we use NME~--- averaged per-vertex Euclidean distances between the recovered and groundtruth meshes, normalized by inter-ocular distances. 
The proposed model has a significantly smaller reconstruction error than the linear model, and is also smaller than the nonlinear model by Tran and Liu~\cite{tran2018on} ($0.0139$ vs.~$0.0146$~\cite{tran2018on}, and $0.0241$~\cite{paysan20093d}).

\begin{figure}[t!]
\begin{center}
\includegraphics[width=0.99\linewidth]{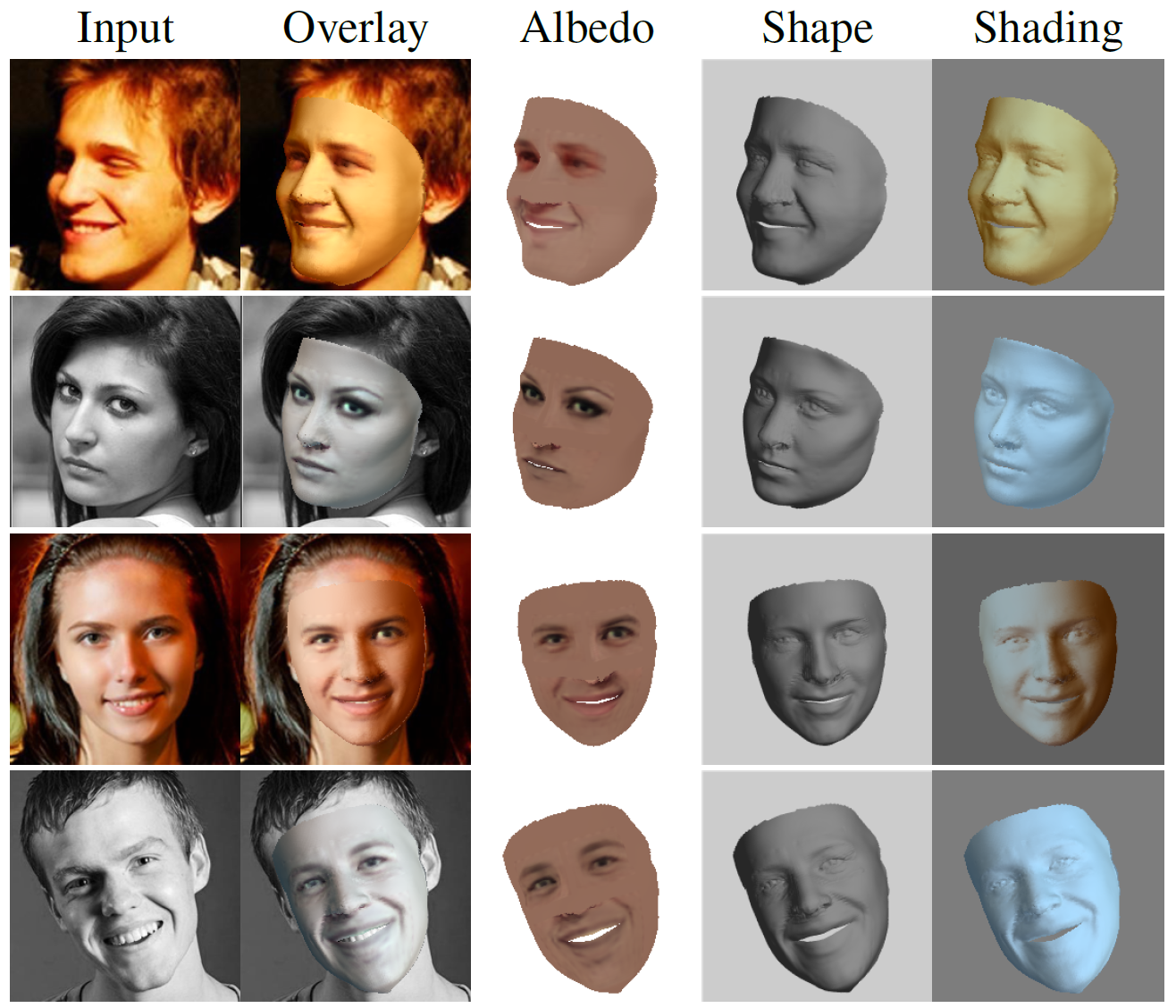}
\vspace{-2mm}
\caption{\small Model fitting on faces with diverse skin color, pose, expression, lighting. Our model faithfully recovers these cues.}
\label{fig:3dmm_fitting}\figvspace 
\end{center}
\end{figure}

\begin{figure}[t!]
\begin{center}
\includegraphics[width=0.99\linewidth]{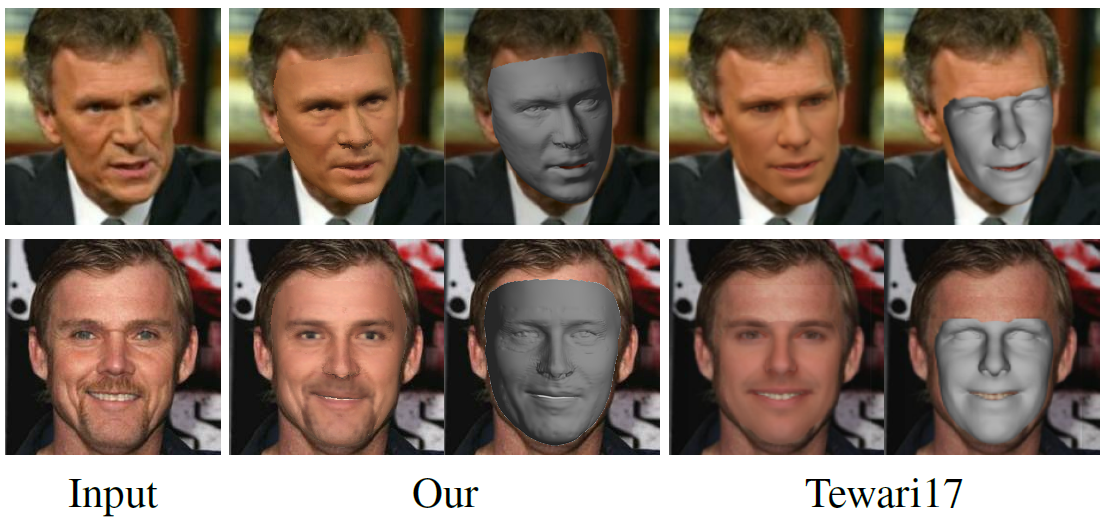}
\vspace{-2mm}
\caption{\small $3$D reconstruction comparison to Tewari~\etal.~\cite{tewari2017mofa}.}
\label{fig:3d_recon_vs_linear}\figvspace 
\end{center}
\end{figure}

\begin{figure*}[t!]
\begin{center}
\includegraphics[width=0.99\linewidth]{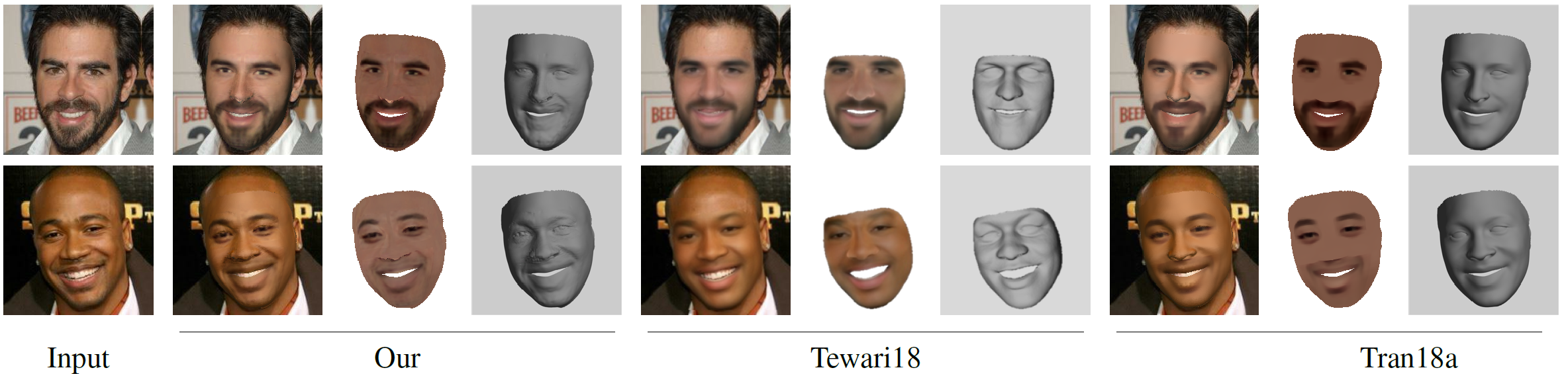}
\vspace{-2mm}
\caption{\small $3$D reconstruction comparisons to nonlinear $3$DMM approaches by Tewari~\etal~\cite{tewari2018self} or Tran and Liu~\cite{tran2018on}. Our model can reconstruct face images with higher level of details. Please zoom-in for more details. Best view electronically.}
\label{fig:3d_recon_vs_nonlinear}\figvspace 
\end{center}
\end{figure*}

\begin{figure}[t!]
\begin{center}
\includegraphics[width=0.99\linewidth]{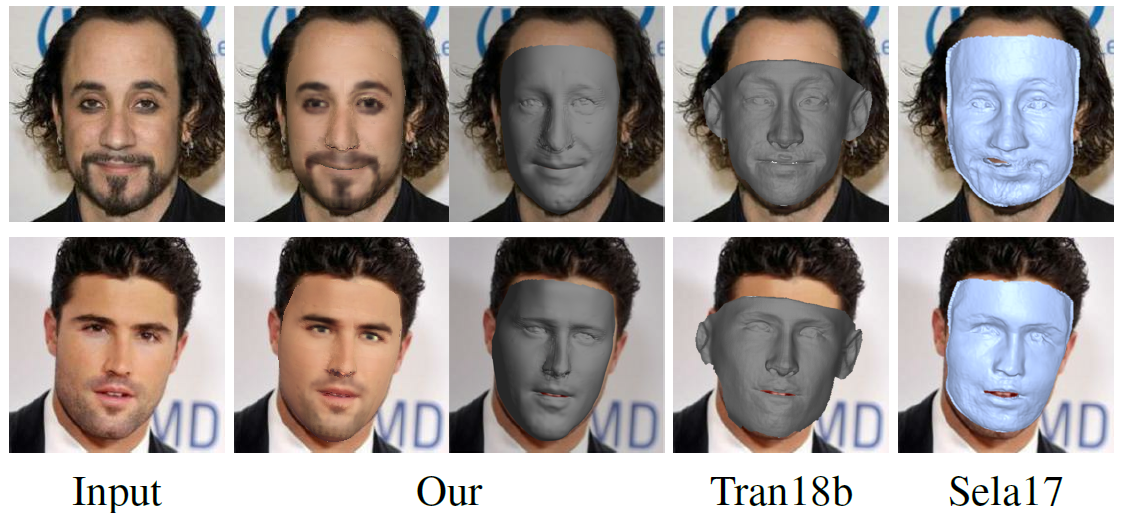}
\vspace{-2mm}
\caption{\small $3$D reconstruction comparisons to Sela~\etal~\cite{sela2017unrestricted} or Tran~\etal~\cite{tran2017extreme}, which go beyond latent space representations. }
\label{fig:3d_recon_vs_non3DMM}\figvspace
\end{center}
\end{figure}

\SubSection{Identity-Preserving}
We explore the effect of our proposed $3$DMM on preserving identity when reconstructing face images. Using DR-GAN~\cite{tran2018representation}, a pretrained face recognition network, we can compute the cosine distance between the input and its reconstruction from different models. Fig.~\ref{fig:id_preserving} shows the plot of these score distributions. At each horizontal mark, there are exactly three points presenting distances between an image with its reconstructions from three models. Images are sorted based on the distance to our reconstruction. For the majority of the cases ($77.2\%$), our reconstruction has the smallest difference to the input in the identity space.

\SubSection{$\mathbf{3}$D Reconstruction}
Using our model $\mathcal{F}_S, \mathcal{F}_A$, together with the model fitting CNN $\mathcal{E}$, we can decompose a $2$D photograph into different components: $3$D shape, albedo and lighting (Fig.~\ref{fig:3dmm_fitting}).
Here we compare our $3$D reconstruction results with different lines of works: linear $3$DMM fitting~\cite{tewari2017mofa}, nonlinear $3$DMM fitting~\cite{tewari2018self, tran2018on} and approaches beyond $3$DMM~\cite{jackson2017large,sela2017unrestricted}. Comparisons are made on CelebA dataset~\cite{liu2015faceattributes}.

For linear $3$DMM model, the representative work, MoFA by Tewari~\etal~\cite{tewari2017mofa, tewari2018high}, learns to regress $3$DMM parameters in an unsupervised fashion.
Even being trained on in-the-wild images, it is still limited to the linear subspace, with limited power to recovering in-the-wild texture. This results in the surface shrinkage when dealing with challenging texture, i.e., facial hair as discussed in ~\cite{tewari2018self, tran2018nonlinear,tran2018on}. Besides, even with regular skin texture their reconstruction is still blurry and has less details compared to ours~(Fig.~\ref{fig:3d_recon_vs_linear}).

The most related work to our proposed model is Tewari~\etal~\cite{tewari2018self}, Tran and Liu~\cite{tran2018on}, in which $3$DMM bases are embedded in neural networks. With more representation power, these models can recover details that the traditional $3$DMM usually can't, i.e. make-up, facial hair. However, the model learning process is attached with strong regularization, which limits their ability to recover high-frequency details of the face. Our propose model enhances the learning process in both learning objective and network architecture to allow higher-fidelity reconstructions~(Fig.~\ref{fig:3d_recon_vs_nonlinear}).

To improve $3$D reconstruction quality, many approaches also try to move beyond the $3$DMM such as Richardson~\etal\cite{ richardson2017learning}, Sela~\etal~\cite{sela2017unrestricted} or Tran~\etal~\cite{tran2017extreme}.
The current state-of-the-art $3$D monocular face reconstruction method by Sela~\etal~\cite{sela2017unrestricted} using a fine detail reconstruction step to help reconstructing high fidelity meshes. However, their first depth map regression step is trained on synthetic data generated by the linear $3$DMM. Besides domain gap between synthetic and real, it faces a more serious problem of lacking facial hair in the low-dimension texture. Hence, this network's output tends to ignore these unexplainable regions, which leads to failure in later steps. Our network is more robust in handling these in-the-wild variations~(Fig.~\ref{fig:3d_recon_vs_non3DMM}). 
The approach of Tran~\etal~\cite{tran2017extreme} shares a similar objective with us to be both robust and maintain high level of details in $3$D reconstruction. However, they use an over-constrained foundation, which loses personal characteristics of the each face mesh. As a result, the $3$D shapes look similar across different subjects~(Fig.~\ref{fig:3d_recon_vs_non3DMM}).

\SubSection{Facial Editting}
\label{sec:relighting}

\begin{figure}[t!]
\begin{center}
\includegraphics[width=0.99\linewidth]{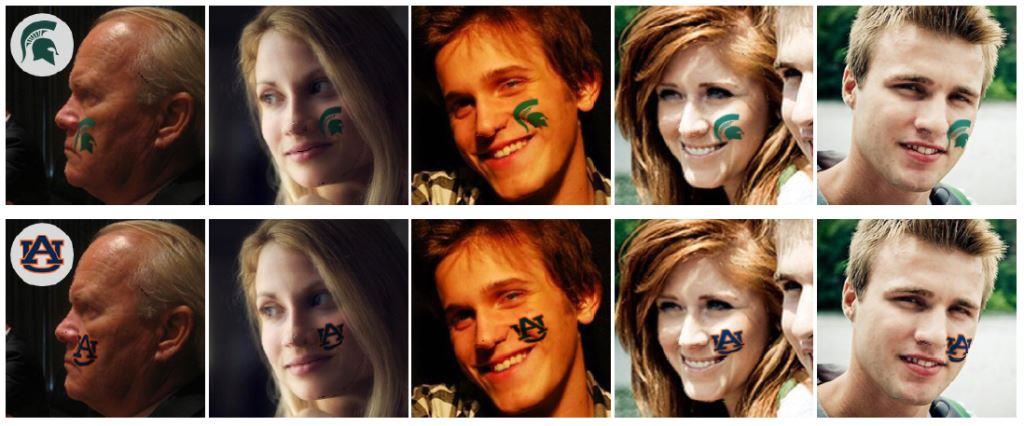}
\vspace{-2mm}
\caption{\small Adding stickers to faces. The sticker is naturally added into faces following the surface normal or lighting.}
\label{fig:editing_sticker}\figvspace
\end{center}
\end{figure}

With more precise $3$D face mesh reconstruction, the quality of successive tasks is also improved. Here, we show an application of our model on face editing: adding stickers or tattoos onto faces. Using the estimated shape as well as the projection matrix, we can unwrap the facial texture into the UV space. Thanks to the lighting decomposition, we can also remove the shading from the texture to get the detailed albedo. From here we can directly edit the albedo by adding sticker, tattoo or make-up. Finally, the edited images can be rendered using the modified albedo together with other original elements.
Fig.~\ref{fig:editing_sticker} shows our editing results by adding stickers into different people's face.

\Section{Conclusions}
\label{sec:con}

In realization that the strong regularization and global-based modeling are the roadblocks to achieve high-fidelity $3$DMM model, this work presents a novel approach to improve the nonlinear $3$DMM modeling in both learning objective and network architecture. Hopefully, with insights and findings discussed in the paper, this work can be a step toward unlocking the possibility to build a model which can capture mid and high-level details in the face. Through which, high-fidelity $3$D face reconstruction can be achieved solely by doing model fitting.

{\small
\bibliographystyle{ieee}
\bibliography{nonlinear_3dmm}
}

\end{document}